\documentclass[11pt,a4paper]{article}
\usepackage[hyperref]{naaclhlt2018}
\usepackage{times}
\usepackage{latexsym}

\usepackage{CJKutf8}
\usepackage{graphicx}
\usepackage{amsmath}
\usepackage{algorithm}
\usepackage{amsfonts}
\usepackage{algorithmic}

\aclfinalcopy 

\title{Distributed Representation for Traditional Chinese Medicine Herb via Deep Learning Models}

\author{Wei Li \\
  MOE Key Laboratory of \\
  Computational Linguistics \\
  Department of Computer Science \\
   Peking University  \\
  {\tt liweitj47@pku.edu.cn} \\\And
  Zheng Yang \\
  School of Chinese Medicine \\
  Beijing University of Chinese Medicine \\
  {\tt yangzheng@bucm.edu.cn}}

\date{}

\begin{document}
\maketitle
\begin{abstract}
Traditional Chinese Medicine (TCM) has accumulated a big amount of precious resource in the long history of development. TCM prescriptions that consist of TCM herbs are an important form of TCM treatment, which are similar to natural language documents, but in a weakly ordered fashion. Directly adapting language modeling style methods to learn the embeddings of the herbs can be problematic as the herbs are not strictly in order, the herbs in the front of the prescription can be connected to the very last ones. In this paper, we propose to represent TCM herbs with distributed representations via Prescription Level Language Modeling (PLLM). In one of our experiments, the correlation between our calculated similarity between medicines and the judgment of professionals achieves a Spearman score of 55.35 indicating a strong correlation, which surpasses human beginners (TCM related field bachelor student) by a big margin (over 10\%). 
\end{abstract}

\section{Introduction}
Traditional Chinese Medicine (TCM) has accumulated a large amount of data during the long term of development, a big part of which embodies as TCM prescriptions. TCM herbs, also known as materia medica, is one of the most important ways of TCM treatment, whose form is the prescriptions that doctor gives based on his/her observation and judgment on the patients' condition. 

The prescriptions consist of various kinds and doses of herbs. We show an example of a famous TCM prescription called Xiao Chai Hu Tang \begin{CJK*}{UTF8}{gbsn}(小柴胡汤)\end{CJK*} in \textbf{Table \ref{TCM prescription example}}. Doctors would adjust the doses of the herbs according to the specific condition of the patient. The herbs have their own natures, for instance, \begin{CJK*}{UTF8}{gbsn}"warm (温)" , "cool (凉)" , "cold (寒)"  and  "hot (热)" .\end{CJK*} Apart from this, the compatibility of medicines also plays a very important role, for example, some certain patterns of combination are strictly prohibited in TCM guidance called "eighteen pairs of strictly prohibited medicine combination \begin{CJK*}{UTF8}{gbsn}(十八反) \end{CJK*}". This indicates that modeling the matching patterns behind the herbs in the prescription is necessary if we want to bring Artificial Intelligence into TCM treatment procedure. 

As the development of data driven kind of machine learning algorithms like deep learning, it has achieved significant improvement in the natural language processing (NLP) field. For instance, neural machine translation \citep{bahdanau2014neural,sutskever2014sequence,DBLP:journals/corr/abs-1710-10393}, text summarization \citep{DBLP:journals/corr/abs-1710-02318,DBLP:conf/acl/MaSXWLS17} question answering \citep{rajpurkar2016squad,wang2016machine}, automatic dialogue generation \citep{li2016deep} and so on. How to apply deep learning to TCM field, which seems relevant to NLP, then becomes an interesting question.

\begin{table}[ht]
\caption{An example of TCM prescription\label{TCM prescription example}}
\begin{center}
\begin{tabular}{|c|p{5cm}|}

\hline
Name & \begin{CJK*}{UTF8}{gbsn}Xiao Chai Hu Tang (小柴胡汤)\end{CJK*}\\
\hline 
Composition & radix bupleuri, Pinellia ternata, ginseng, licorice root, Scutellaria baicalensis, ginger, Chinese-date \\ \hline
Translation & \begin{CJK*}{UTF8}{gbsn}柴胡、半夏、人参、甘草、黄芩、生姜、大枣。\end{CJK*}\\
\hline

\end{tabular}
\end{center}
\end{table}  

Previous works have attempted to use probabilistic topic models such as Latent Dirichlet Allocation (LDA) and Latent Semantic Analysis (LSA) to describe the properties of the herbs \citep{Zhang2011Topic,Zhu2017TCM}. In NLP field, the neural network based word embedding models \citep{mikolov2013distributed,pennington2014glove} have achieved great success, and laid a good foundation for the development and application of deep learning models \citep{collobert2011natural}. In this paper, we propose to learn the distributed representation of TCM herbs by a way analogous to the word embeddings in NLP, which can be hopefully helpful to the further development of TCM research. 

However, TCM prescriptions are not exactly like natural language sentences. TCM prescriptions have their own way of organizing the herbs, which are often put in a weakly ordered way. The herbs in the front of the prescription may be connected with the very last herb instead of the surrounding ones. In our learning process, We see each prescription as a sequence of tokens. The herbs form the context to each other. By predicting the central herb with the corresponding context, we can learn the representation of each herb, which contains the information of the patterns of combination, indicating some of the properties of the herbs. In our experiments we see that first modeling the prescription as a whole provides much better results than traditional language modeling style methods.

Although there has been thousands of TCM prescriptions in the history, because of the lack of digitalization, there has not been much structured digital resources. In this paper, we collect large scale digital resources from the Internet. After some steps of formalization and cleaning of the data, we get over 80,000 TCM prescriptions. By predicting the randomly chosen central herb based on the corresponding context, we can learn the representation of each herb, which contains the information of the patterns of combination. In this paper we propose a Prescription Level Language Modeling (PLLM) that predicts the central herb by first modeling the whole prescription. In our experiments we observe that our PLLM method performs much better than traditional language modeling style methods. Apart from these, we also propose one possible way of applying deep learning to assist doctors in TCM treatment in the real life.

Our contributions mainly lie in the following aspects:
\begin{itemize}
\item We clean and formalize a large scale of TCM data from the Internet and provide a dataset for training and testing the quality of herb embeddings. 
\item We propose to represent TCM herbs with distributed embeddings, and propose a Prescription Level Language Modeling (PLLM) method to learn the distributed representations of the TCM herbs. In the experiments we see that modeling the prescription as a whole is better than directly applying language modeling method.
\item We propose a possible way to assist TCM doctors to compose prescriptions with deep learning methods.
\end{itemize}

\section{Related Work}
\subsection{Computational TCM Methods}
\citet{zhou2010development} attempted to build TCM clinical data warehouse to make use of TCM knowledge. This is a typical way of collecting big data, since the number of prescriptions given by the doctors in the clinics is very large. However, in reality, besides the problem of quality, most of the TCM doctors don't use these digital systems. Therefore, we choose prescriptions in the traditional classics of TCM. Although this may suffer from the loss of data magnitude, it guarantees the quality of the prescriptions. 

\citet{wang2004self} attempted to construct a self-learning expert system with several simple classifiers to facilitate the TCM diagnosis procedure, \citet{Wang2013TCM} proposed to use deep learning and CRF based multi-labeling learning methods to model TCM inquiry process, but their systems are too simple to be actually effective in the real life TCM diagnosis. \citet{lukman2007computational} made a survey on some computational methods on TCM, while these methods utilize traditional data mining methods.

\subsection{Distributed Word Embedding}
\citet{bengio2003neural} first proposed to learn the distributed representation of words while predicting the next word in a sequence to fight the curse of dimensionality. \citet{mikolov2010recurrent} followed this thread by expanding the simple feed forward neural networks to recurrent neural networks, hoping to capture longer distance dependency. These two models still largely resembles the framework of probabilistic language modeling. 

\citet{mikolov2013distributed,mikolov2013efficient} proposed two very simple yet effective models called continuous bag of words (CBOW) and skip gram. CBOW predicts the central word in a context window based on the words in the window with a simple logistic regression classifier. Skip gram uses the same architecture but predicts the context words based on the central word. Although these two models achieved very good results in many kinds of tasks, they suffer from the loss of not being able to utilize global information. To tackle this problem, \citet{pennington2014glove} proposed a Global Vector model (GloVe), which aims to combine the advantage of both LSA model and the CBOW model. We develop our methods to learn the distributed representations of herbs inspired by the above ideas while modeling the prescription as a whole rather than using limited context window.

\section{Data Construction}
When constructing our TCM prescription dataset, we first considered the TCM medical records \begin{CJK*}{UTF8}{gbsn}(中医医案)\end{CJK*} in the history, which contain a lot of very good resource. The medical records are widely referenced by the doctors in the treatment, however, they have not been well digitalized, which makes it hard to extract the prescriptions out of the descriptive natural language from the records. Another way to get large scale prescriptions is from TCM clinics, the problem is that most of this kind of valuable data is not publicly available. Therefore, we turn to Internet resources, which contain large scale digitalized prescription resources. 

We crawl the data from \begin{CJK*}{UTF8}{gbsn}TCM Prescription Knowledge Base\  (中医方剂知识库)\end{CJK*} \footnote{\url{http://www.hhjfsl.com/fang/}}. This knowledge base includes quite comprehensive TCM documentations in the history, which also provides a search engine for prescriptions. The database includes 710 TCM historical books or documentations as well as some modern ones, consisting of 85,166 prescriptions in total. Each item in the database  provides the name, origin, compositions, effect,  prescription, contraindications and preparation methods. We clean and formalize the database and get 85,161 usable prescriptions\footnote{The data and processing code are all available on-line}. 

In the process of normalization, we temporarily omit the dose information and the preparation method description, which we may use in the future.
Word segmentation is typically the first step to Chinese text processing \citep{DBLP:conf/acl/XuS16,journals/talip/ZhaoHLL10,SunLWL14,DBLP:conf/acl/SunWL12,DBLP:conf/naacl/SunZMTT09}. Word segmentation is used to pre-process the text into word based sequences. In addition to the traditional word segmentation techniques, we use more heuristics to assist the segmentation process because this domain has specific features.
We also write some simple rules to project some rarely seen herbs to their similar form that is normally referred to. For example, if the herb appears less than 5 times and all the  characters of the herb name is a substring of another more popular herb, then the herb would be mapped to the other one. This simple projection procedure can partly solve the data sparsity problem.

\subsection{HerbSim80\label{HerbSim80}}
Similar to the way of building wordsim353 \citep{finkelstein2001placing} , we manually build a dataset consisting of 80 pairs of herbs, which we ask three TCM professionals to make a judgment on how likely the two herbs in the pair would appear in the same prescription. We then evaluate the embeddings by calculating the correlation between the similarity scores given by the cosine distance of embeddings and the scores given by the professionals.
In Table \ref{HerbSim80 Example} we show some examples. 
\begin{table}[ht]
\begin{center}
\caption{Examples of HerbSim80\label{HerbSim80 Example}. The Chinese characters in Herb 1 and Herb 2 two columns are all herb names. The scores in S 1, S 2 and S 3 are scores given by the three professionals. Ave S is the average of three scores.\footnote{\begin{CJK*}{UTF8}{gbsn}乌头\end{CJK*}}}
\begin{tabular}{|c|c|c|c|c|c|}
\hline
Herb 1 & Herb 2 & S 1 & S 2 & S 3 & Ave S \\ \hline

\begin{CJK*}{UTF8}{gbsn}乌头\end{CJK*} & \begin{CJK*}{UTF8}{gbsn}栀子\end{CJK*} & 1 & 1 & 1& 1.00 \\
\begin{CJK*}{UTF8}{gbsn}麦门冬\end{CJK*} & \begin{CJK*}{UTF8}{gbsn}山茱萸\end{CJK*} & 3 & 3 & 3 & 3.00\\
\begin{CJK*}{UTF8}{gbsn}赤芍\end{CJK*} & \begin{CJK*}{UTF8}{gbsn}藿香\end{CJK*} & 2 & 2 & 1 & 1.67 \\
\begin{CJK*}{UTF8}{gbsn}苍术\end{CJK*} & \begin{CJK*}{UTF8}{gbsn}乌梅肉\end{CJK*} & 2 & 1 & 1 & 1.33 \\ \hline

\end{tabular}
\end{center}
\end{table}

The detailed procedure is as follows:
\begin{enumerate}
\item We randomly generate 120 pairs of TCM herbs. 
\item We invite three TCM professionals, who have been practicing TCM diagnosis and treatment for over five years, to give a score of the herb pair between 1 and 5. 1 indicates that the two herbs are very unlikely to appear together in one prescription. On the contrary, 5 indicates that the two herbs often appear as a pair in the same prescription. 
\item We rank the pairs by the standard deviation between the three scores given by the professionals, and get the top 80 pairs with better agreement. The final score is set to be the average of the three scores.
\item We invite a junior student who majors in TCM (student who has just finished the course of Principles of TCM Prescriptions) to do the task, which will be compared with the result given by the embeddings.
\end{enumerate}

\section{Distributed Representations of TCM Herb\label{Models}}
Similar to the one-hot representation in NLP, herbs can also be represented as one-hot vectors, where the length of the vector is the size of the whole herb vocabulary and in each vector, only one slot is filled with ``1'' while others are all ``0''. The problem of this way of representation is that it can not show the innate relation between herbs, which is even more important than it is in NLP. For example, cinnamon\begin{CJK*}{UTF8}{gbsn}(肉桂)\end{CJK*} and cinnamon twig \begin{CJK*}{UTF8}{gbsn}(桂枝)\end{CJK*} are two different parts of the same plant cinnamon tree. The natures of these two herbs are very similar, but in the one-hot style representation, the distance between these two herbs are no different from any other pairs.

Another possible way of representing herbs is to model the herbs with some features that how TCM experts view the herbs. For each aspect of the herb, we can use one-hot vectors to represent them. This way of representation accords to the theory of TCM research in the history. For example, assume we model each herb from \begin{CJK*}{UTF8}{gbsn}cold and hot\ (寒热)、the five flavors\ (五味\ : sweet, sour, bitter, pungent and salty)\end{CJK*} two aspects, we can represent a herb with a 7-length vector. However, to make this work, it costs very expensive human expert effort, which makes it impracticable.

Inspired by the way of representing words with distributed vectors in NLP field, we propose to represent the TCM herbs with distributed representations. We model the TCM prescriptions as documents in NLP, while herbs as words. We can automatically learn the herb embeddings by tuning the distributed representation of herbs while predicting the central herb with context herbs in the prescription. This way the information of the herb is implicitly embodied in the vectors, and we can learn the representations automatically from the dataset we build without much human effort. 

Although TCM prescriptions are very similar to natural language texts, there is one major difference that in natural language, the order of the words is very important which is strictly restricted by syntax and grammar, while in prescriptions, the order of the herbs usually plays less important roles. On the other hand, the herbs in the front of a prescription may be connected to the very last one instead of its surrounding ones. Based on this observation, we propose to first model the prescription as a whole, and then predict the central herb. 

\subsection{Proposed Baselines}
In this subsection, we propose several baseline models that are directly adapted from NLP field.
\begin{itemize}
\item Latent Semantic Analysis (LSA) : LSA is used to analyze the relationship between a set of documents and the terms they contain by producing a set of concepts related to the documents and terms. We model the herbs as words and prescriptions as documents. We use a  matrix containing herb counts in prescriptions to represent the coexistence relation between herbs and prescriptions. With singular value decomposition, we can get a vector for each herb, which is then used as the distributed representation of the herbs.
\item Continuous bag of words (CBOW)\citep{mikolov2013distributed} : This model uses a log-linear regression model with negative sampling to predict a central word with the context words within a local context window.
\item Global vectors for word representation (GloVe)\citep{pennington2014glove} : This model uses a global log-bilinear regression model that combines the advantages of the global matrix factorization (similar to LSA) and local context window (similar to CBOW) methods.
\item Recurrent Neural Networks Language Modeling (RNNLM) : This model is similar to CBOW model in the setting of objective, that is to predict a herb based on their context herbs. However, this model aims to model a longer dependency between herbs by using a bidirectional gated recurrent neural networks (BiGRNN) to predict the central herb. This model predicts the central herb by considering the herbs both before and after it.

\end{itemize}
\subsection{Prescription Level Language Modeling}
In this subsection, We show the details of our proposed methods, Prescription Level Language Modeling(PLLM).

\begin{figure}[ht]
\centering
\includegraphics[scale=0.3]{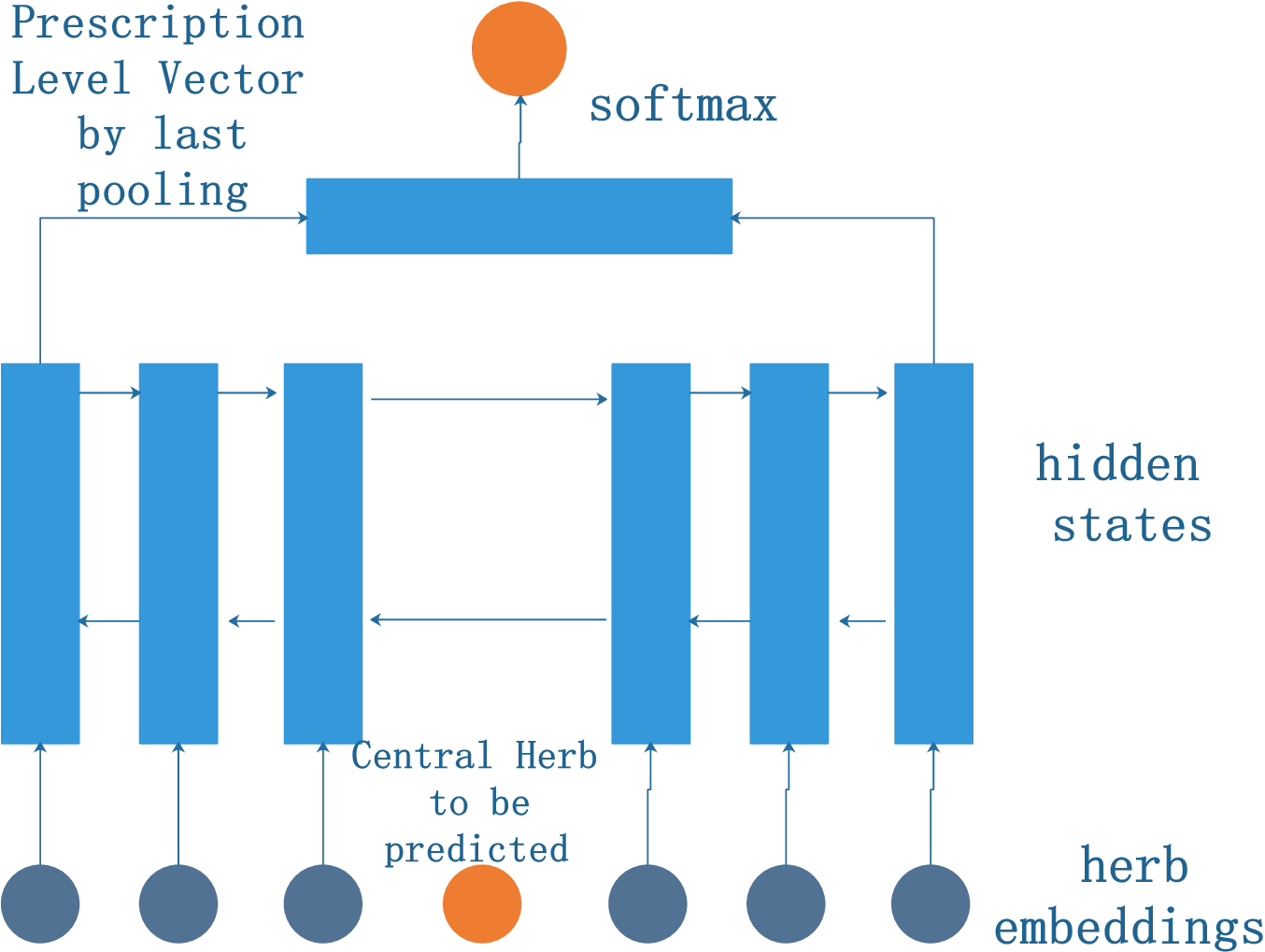}
\caption{An illustration of Prescription Level Language Modeling with BiGRNN\label{GRNN}}
\end{figure}

In this subsection, We show the details of our proposed Prescription Level Language Modeling. As is shown in Figure \ref{GRNN}, we first encode the whole prescription (except the central herb) into a prescription level vector (vector that encodes the information of the whole prescription) $h_{c}$, and then predict the central herb based on this prescription level vector $h_{c}$. 

\begin{enumerate}
\item We take the one-hot herbs in the prescription $w_{1}, ..., w_{t-1}, w_{t+1}, ... w_{n}$ as input, and project them into the corresponding embeddings $e_{1},...e_{t-1},e_{t+1},...,e_{n}$.
\item Then we go over the whole prescription except the central word to be predicted $w_{t}$ with BiGRNN, which gives us the hidden states $h_{1},...,h_{t-1},h_{t+1},...,h_{n}$.
\item After that, We apply last pooling with the hidden states, and get the context vector $h_{c}$, which is expected to embody the information of the whole prescription.
\item Finally, we use a $softmax$ regression layer to predict the central herb.
\end{enumerate}

We encode the whole prescription into a fixed length vector $h_c$ in order to capture the dependency beyond the local windows of the herbs. In this way, even the last herb can be an auxiliary to the first herb in the prescription. We believe this is very important compared with the baseline RNNLM model, as it separates the herbs before and after the central one. Further more, the vector of the whole prescription can also be a good representation of the disease that the prescription wants to tackle, which we would like to explore in the future.

\section{Application on TCM Treatment}
\label{sec:Application}
In the TCM diagnosis and treatment procedure, different from modern medicine science, doctors usually have more freedom when writing a prescription based on his own observation instead of a standard process. Still, they often refer to the classical prescriptions recorded in the TCM classics, for instance, \textit{Treatise on Febrile Diseases} (\begin{CJK*}{UTF8}{gbsn}《伤寒论》\end{CJK*}). In these classics, there are not only the principles for giving the prescriptions but also some widely used, carefully constructed prescriptions. In this section, based on this observation, we propose a language modeling style method based on the model we learn from the classical prescriptions, which can hopefully give hints to doctors on writing prescriptions for patients. On thing that should be noted is that our proposed method is more of a prototype rather than a complete tool.

Doctors start to write prescriptions after they have made a judgment on the patients' situations. Each time a herb is given, our model would process the unfinished prescription, and then suggest a candidate herb that the doctor may want to use. The CBOW, RNNLM, PLLM models are the same as described in Section \ref{Models}. We also apply N-gram model as our baseline model. N-gram model is similar to how it is used in NLP, which predicts the next herb by selecting the herb with the largest likelihood. The likelihood is given by the linear combination of unigram, bigram and trigram transition probabilities. The parameters of these models are all trained on the dataset we build, which consists of classical prescriptions. After our model predicts the most probable herb, doctors can choose whether to take the advice or not.

\begin{algorithm}
\caption{TCM prescription auxiliary composition}
\begin{algorithmic}[1]
\REQUIRE The unfinished prescription, trained model
\STATE Read in the herbs in the unfinished prescription
\STATE Choose a model out of n-gram, RNNLM, CBOW, PLLM to process the herbs.
\STATE Predict a herb that fits the current unfinished prescription.
\end{algorithmic}
\end{algorithm}

\section{Experiments}
In all of the following experiments, we use our distributed herb representations in an unsupervised way. The distance between two herbs are all given by the cosine distance between the two vectors.
\subsection{Correlation}
In this section, we show the correlation results between the professionals and various models. We use the \textbf{HerbSim80} dataset described in Section \ref{HerbSim80}. For LSA, the vector size is set to be 20, while for other models, the vector size is set to be 100. The gensim toolkit \footnote{\url{http://radimrehurek.com/gensim/}} is used to train the LSA model. For GloVe\footnote{\url{https://nlp.stanford.edu/projects/glove/}} and CBOW\footnote{\url{https://code.google.com/p/word2vec/}} we use their official program respectfully. The similarity score of two herbs is given by the cosine distance between the vectors of the herbs. We use Spearman's rank score as the criteria to evaluate the correlation between our model and the professionals. Our RNNLM, PLLM models are built using Tensorflow toolkit\citep{tensorflow2015-whitepaper}. We choose Adam\citep{kingma2014adam} as the optimization method. An early stopping strategy is adopted to avoid over-fitting in the training process. We stop the training process when the accuracy of herb prediction in the development set fails to increase in the last three training epochs.

In the bottom row of Table \ref{Correlation}, we show the correlation result of a human junior student who majors in TCM. From the table we can see that PLLM model gives the best result, which surpasses the result of the student by over 10\%. This phenomenon shows that our PLLM model can learn some useful knowledge out of the prescriptions in the dataset with unsupervised learning. An overall description of the prescription can indeed help predicting the herb. The simple model CBOW also gives a rather good result of 49.33\%.  The traditional LSA model doesn't perform well in this experiment, maybe because it omits the local information of the herbs, which plays a more important role in TCM prescriptions. GloVe suffers the same problem that the objective of global part influences the representation of the local context. 

\begin{table}[ht]
\caption{Correlation with Professionals' judgment\label{Correlation}.  $Student$ means the correlation score with the Junior student.}
\centering
\begin{tabular}{|c|c|}
\hline
Model & Spearman's R \\ \hline
Student & 0.4506 \\ \hline
LSA & 0.3861 \\
GloVe & 0.3952 \\
CBOW & 0.4933 \\
RNNLM & 0.5158 \\ \hline
\textbf{PLLM} & \textbf{0.5535} \\ \hline

\end{tabular}
\end{table}

\subsection{Prediction}
In the Section \ref{sec:Application}, we propose to use our model to assist doctors to write prescriptions. We manually build a test set consisting of 206 prescriptions. For each prescription, we temporarily blank one of the herbs, which is randomly chosen, and test whether our models could predict  what the original herb is. The original prescriptions all have at least four herbs. Some examples are shown in Table \ref{Prediction Example}, where the herb in the Answer column is the blank in the Question column.

\begin{table}[ht]
\caption{Examples from the test set for herb prediction\label{Prediction Example}. The characters in the \textbf{Question} column are TCM herb names. The characters in the \textbf{Answer} column is the herb name that should be put into the blank in the Question}
\begin{center}
\begin{tabular}{|c|c|}
\hline
Question & Answer \\ \hline
\begin{CJK*}{UTF8}{gbsn}麻黄\ \_\_\_ 杏仁 \  炙甘草\end{CJK*} & \begin{CJK*}{UTF8}{gbsn}桂枝\end{CJK*} \\
\begin{CJK*}{UTF8}{gbsn}生地黄\ 当归\ 牡丹皮\  \_\_\_ 升麻\end{CJK*} & \begin{CJK*}{UTF8}{gbsn}黄连\end{CJK*} \\
\hline
\end{tabular}
\end{center}
\end{table}

In this experiment, we use bigram and trigram from both directions for n-gram prediction. For the prediction score, we simply add up the probabilities with the same weights.
\begin{equation}
\begin{split}
score(i) = p(i|i-1) + p(i|i-2,i-1)\\+ p(i|i+1) + p(i|i+1,i+2)
\end{split}
\end{equation} 

From Table \ref{Acc prediction} we can see that the baseline model N-gram is very strong. The accuracy of N-gram model is even higher than RNNLM model, which shows that directly transfer the language modeling method from NLP may not be a good idea when predicting the next herb. CBOW model is slightly different from the original one, we average all the herb embeddings of the prescription, based on which we predict the blanked one. We assume the reason that it gives a rather good result is that it makes use of a wider range of context.  In this experiment, our PLLM gets the best result, which is much higher than other models. We assume that it is necessary to consider the whole prescription when predicting the next herb.

Again we clarify that this application is a prototype. It doesn't mean that we don't need to consider other factors like the patients' situation when composing a prescription. What we want to show is that the combination of herbs play an important role when composing a prescription and our model can capture this kind of pattern on some level.

\begin{table}[ht]
\caption{Accuracy on herb prediction}\label{Acc prediction}
\centering
\begin{tabular}{|c|c|}
\hline
Model & Accuracy \\ \hline
N-gram & 17.96 \\ 
CBOW & 20.39 \\
RNNLM & 16.50 \\ \hline
\textbf{PLLM} & \textbf{32.04} \\ \hline
\end{tabular}
\end{table}

\subsection{Further Observation}
In the experiments, we observe that the distributed vectors of herbs have linear algebraic relationships. For example, $v$[\begin{CJK*}{UTF8}{gbsn}熟地黄 (prepared rehmannia root)\end{CJK*}] $-\ v$[\begin{CJK*}{UTF8}{gbsn}生地黄 (dried rehamnnia root)\end{CJK*}] $\approx v$ [\begin{CJK*}{UTF8}{gbsn}煨姜 (roasted ginger)\end{CJK*}] $-\ v$ [\begin{CJK*}{UTF8}{gbsn}生姜 (ginger)\end{CJK*}]. This phenomenon is similar to the observation described in \cite{mikolov2013efficient}, where $v[bigger] - v[big] \approx v[smaller] - v[small]$. In the future, we hope to further look into this and see whether this is a general phenomenon in TCM herbs.

\section{Conclusion and Future Work}
In this paper, we propose to represent TCM herbs with distributed representation via Prescription Level Language modeling. In the experiments we testify that simply adopting the methods from NLP field is problematic because of the difference that lies between natural language and TCM prescriptions. Furthermore, we propose a possible application for our models in TCM treatment, which we hope can facilitate doctors composing prescriptions.


\bibliography{naaclhlt2018}
\bibliographystyle{acl_natbib}

\appendix

\end{document}